# Improving Character-level Japanese-Chinese Neural Machine Translation with Radicals as an Additional Input Feature

Jinyi Zhang
Graduate School of Engineering
Gifu University
Gifu, Japan
e-mail: zhang@mat.info.gifu-u.ac.jp

Tadahiro Matsumoto
Faculty of Engineering
Gifu University
Gifu, Japan
e-mail: tad@gifu-u.ac.jp

*Abstract*—In recent years, Neural Machine Translation (NMT) has been proven to get impressive results. While some additional linguistic features of input words improve word-level NMT, any additional character features have not been used to improve character-level NMT so far. In this paper, we show that the radicals of Chinese characters (or kanji), as a character feature information, can be easily provide further improvements in the character-level NMT. In experiments on WAT2016 Japanese-Chinese scientific paper excerpt corpus (ASPEC-JP), we find that the proposed method improves the translation quality according to two aspects: perplexity and BLEU. The character-level NMT with the radical input feature's model got a state-of-the-art result of 40.61 BLEU points in the test set, which is an improvement of about 8.6 BLEU points over the best system on the WAT2016 Japanese-to-Chinese translation subtask with ASPEC-JP. The improvements over the character-level NMT with no additional input feature are up to about 1.5 and 1.4 BLEU points in the development-test set and the test set of the corpus, respectively.

*Keywords-neural machine translation; character-level; Japanese-Chinese; character feature; radical*

## I. INTRODUCTION

Neural Machine Translation (NMT) [1], [2] has made remarkable achievements recently. Most existing machine translation (MT) systems operate at the level of words, relying on explicit segmentation to extract tokens. One of the weaknesses of the word-level NMT is limitation of vocabulary size due to its architecture. In addition, obtaining uniformed correct word segmentation results is not straightforward for some languages such as Chinese and Japanese, since the words are not separated in their written form.

The characters used in a language are much fewer than the words of the language. Character-level neural language models [3] and MT are explored and achieved respective results. Previous works, such as POS tagging [5], name entity recognition [6], parsing [7], and learning word representations [8], show different advantages of using character-level information in Natural Language Processing (NLP).

Besides, subword-based representations (the middle of word-based and character-based representations) have been explored in NMT [4], and are applied to English and other western languages, where almost all words consist of several characters. Contrastingly, Chinese characters, which are used in written Chinese, Japanese and some other Asian languages, are typical logograms. A logogram is a character that represents a concept or thing, namely a word; and thus, it is difficult to split those words into subwords.

Sennrich et al. [9] show that linguistic input features, such as POS tags and lemmas, improve word-level NMT. We conjecture that some feature information of the input characters also could be valuable for character-level NMT. In this paper, we investigate whether radicals of Chinese characters, as character feature information, are beneficial to character-level NMT. Li et al. [10] prove radicals are useful in other NLP applications.

Unlike the alphabets of European languages, Chinese characters are logograms, of which over 80% are phono-semantic compounds, with a semantic component giving a broad category of meaning and a phonetic component suggesting the sound. For example, the radical 金 (metal) of the Chinese character 鉄 (iron) in Japanese provides the meaning connected with metal. In fact, the radical of most Chinese characters inherently bring with certain levels of semantics regardless of the contexts. Being aware that the radical of Chinese characters are finer grained semantic units, so the radical as the input feature information will convey the meaning of the Chinese characters, and improve the translation performance.

In this paper, we follow the WAT2016 [11] published system as the baseline system, and use [2] as the base NMT system, which follows an encoder-decoder architecture with attention in word-level. In our case, we take advantage of Chinese character information in the character-level NMT.

In the following sections, Section 2 briefly explains the architecture of the NMT that we are using as the base system. Section 3 describes the proposed method, input feature information for Japanese. Section 4 reports the experimental framework and the results obtained in the Japanese-to-Chinese WAT2016 subtask (with ASPEC-JC [12]) that improve the translation performance. Finally, section 5 concludes with the contributions of the paper and further work.

## II. NEURAL MACHINE TRANSLATION

NMT uses a neural network approach to compute the conditional probability $p(y|x)$ of the target sentence $y$ for a given source sentence $x$. We follow the NMT architecture by Luong et al. [2], which we briefly describe here with reference to [9]. The NMT system is implemented as a global attentional encoder-decoder network with recurrent neural networks, and we simply use it in character-level.

First, the encoder is a bidirectional neural network with Long Short-Term Memory (LSTM) units that reads an input sequence $x = (x_1, ..., x_m)$ and calculates a forward sequence of hidden states $(\vec{h}_1, ..., \vec{h}_m)$ and a backward sequence $(\overleftarrow{h}_1, ..., \overleftarrow{h}_m)$. Then, the hidden states $\vec{h}_j$ and $\overleftarrow{h}_j$ are concatenated to obtain the annotation vector $h_j$. The decoder is a recurrent neural network with LSTM units that predicts a target sequence $y = (y_1, ..., y_n)$. Thirdly, each word (or character in case of character-level NMT) $y_i$ is predicted based on a recurrent hidden state $s_i$, the previously predicted word (or character) $y_{i-1}$, and a context vector $c_i$. $c_i$ is computed as a weighted sum of the annotations $h_j$. Finally, the weight of each annotation $h_j$ is computed through an alignment (or attention) model $\alpha_{ij}$, which models the probability that $y_i$ is aligned to $x_j$.

The forward states of the encoder is expressed as below:

$$\vec{h}_j = \tanh(\vec{W} E x_j + \vec{U} \vec{h}_{j-1}) \quad (1)$$

where $E \in \mathcal{R}^{p \times V_x}$ is a word embedding matrix, $\vec{W} \in \mathcal{R}^{q \times p}$ and $\vec{U} \in \mathcal{R}^{q \times q}$ are weight matrices; $p$, $q$ and $V_x$ are the word embedding size, the number of hidden units, and the vocabulary size of the source language, respectively.

Equation (1) can be generalized to an arbitrary number of features $|F|$ as follows:

$$\vec{h}_j = \tanh(\vec{W} (\|_{k=1}^{|F|} E_k x_{jk} + \vec{U} \vec{h}_{j-1})) \quad (2)$$

where $\|$ is the vector concatenation, $E_k \in \mathbb{R}^{p_k \times V_k}$ are the feature embedding matrices, with $\sum_{k=1}^{|F|} p_k = p$, and $V_k$ is the vocabulary size of the *k*th feature. The embedding of the input features is trained separately, in the same way as the word embedding. Finally embedding of the input features concatenated with the word embedding. The length of the concatenated vector matches the total embedding size.

### III. INPUT FEATURE FOR JAPANESE CHARACTERS

A radical (Chinese: 部首; pinyin: bùshǒu; literally: "section header") is a graphical component of a Chinese character under which the character is traditionally listed in a Chinese dictionary.

The English term "radical" is based on an analogy between the structure of characters and inflection of words in European languages. Radicals are also sometimes called "classifiers," but this name is more commonly applied to grammatical classifiers (measure words). Normally, the radical is also the semantic component.

Here we used the 214 Kangxi radicals (康熙部首) partially shown in Table I, which form a system of radicals of Chinese characters. The radicals are numbered in stroke count order. They are the de facto standard used as the basis for most modern Chinese dictionaries. They serve as the basis for many computer encoding systems. Specifically, the Unicode standard's radical-stroke charts are based on the Kangxi radicals or radicals.

The modern Japanese is written in three main writing systems: kanji (漢字), characters of Chinese origin used to represent both Chinese loanwords into Japanese and a number of native Japanese morphemes; and two syllabaries: hiragana and katakana. Kanji radicals are graphemes, or graphical parts, that are used in organizing Japanese kanji in dictionaries. They are derived from the 214 Chinese Kangxi radicals.

Hiragana are used for words without kanji representation such as particles (e.g. が, を), for words no longer written in kanji, and also following kanji to show conjugational endings (e.g. 書く). Katakana are primarily used to write foreign words, plant and animal names, and for emphasis.

There is no radical in hiragana and katakana characters themselves. However, they are derived from kanji, which have radicals. For this reason, we borrow the radicals of the original kanji of hiragana and katakana characters for them. Fig. 1 shows the derivation of hiragana from manyōgana (an ancient writing system that employs Chinese characters to represent the Japanese language) via cursive script. The upper part shows the character in the regular script form, the middle part shows the cursive script form of the character, and the bottom shows the equivalent hiragana. From the upper part, you can find a hiragana's original kanji, and then get the radical of the original kanji. You can obtain the radicals of katakana from the original kanji in the same way.

The input Japanese sentences in the parallel training data contain not only the Japanese characters but also Arabic numerals, English alphabet and various symbols. In order to make every input character have its radical (from the 214 Chinese Kangxi radicals) as its feature, we assign radicals to those non-Japanese characters as follows.

- For Arabic numerals 0, 1, 2, 3, ..., 9, we borrow the radicals of the corresponding Chinese numerals (零, 一, 二, 三, ..., 九), i.e., 雨, 一, 二, 一, ..., 乚.
- For English alphabet, we use 英's radical ⺾ equally (英 represents English or Britain in Japanese).
- For various symbols, we use 符's radical ⺮ equally (符号 represents symbols in Japanese).

TABLE I.  EXAMPLES OF 214 KANGXI RADICALS

| 一 | 丨 | 丶 | 丿 | 乙 | 亅 | 二 | 亠 | 人 | 儿 | 入 | 八 | 冂 | 冖 | 冫 | 几 |
|---|---|---|---|---|---|---|---|---|---|---|---|---|---|---|---|
| 士 | 夂 | 夊 | 夕 | 大 | 女 | 子 | 宀 | 寸 | 小 | 尢 | 尸 | 屮 | 山 | 巛 | 工 |

Figure 1. Derivation of hiragana
(https://en.wikipedia.org/wiki/Hiragana)

We employed the Python library cjklib[1] to obtain the radical from a kanji character (we slightly modified the software to enable it in Python3 environment for 8-bit Unicode Transformation Format).

Finally, we got the input features that only consist of the 214 Chinese Kangxi radicals.

Table II shows a part of a Japanese source sentence and the corresponding Chinese Kangxi radicals. This sentence includes Japanese (kanji and hiragana) characters, Arabic numerals, English alphabet, and a symbol.

## IV. EVALUATION AND RESULTS

We evaluate our system on the ASPEC-JC Japanese-Chinese corpus, which were shared for the WAT2016 Japanese-to-Chinese translation subtask. This corpus was constructed by manually translating Japanese scientific papers into Chinese [11], [12]. The Japanese scientific papers are either the property of the Japan Science and Technology Agency (JST) or stored in Japan's Largest Electronic Journal Platform for Academic Societies (J-STAGE). ASPEC-JC is composed of 4 parts: training data(672,315 sentence pairs), development data(2,090 sentence pairs), development-test data(2,148 sentence pairs) and test data(2,107 sentence pairs) on the assumption that it would be used for machine translation research. ASPEC-JC contains both abstracts and some parts of the body texts. ASPEC-JC only includes "Medicine", "Information", "Biology", "Environmentology", "Chemistry", "Materials", "Agriculture" and "Energy" fields because it was difficult to include all the scientific fields. These fields were selected by investigating the important scientific fields in China and the use tendency of literature database by researchers and engineers in Japan.

Our LSTM models have 1 layer, each with 512 cells, and embedding size is 512. The parameters are uniformly initialized in [−0.1, 0.1], using plain SGD, starting with a learning rate of 1. The mini-batch size is 10. The normalized gradient is rescaled whenever its norm exceeds 1. The dropout probability is set to 0.8. Decoding is performed with beam search with a beam size of 5. To calculate the BLEU score, we divided the output sentences into words by Jieba[2], a Python module for Chinese text segmentation. We also used the input-feeding approach by Luong et al. [2], which had been proved better performance in the model.

Our code is implemented on OpenNMT [13]. The speed of the training was 3K target characters per second on a single GPU device, GeForce GTX 1080. It takes 3–4 days to train a model completely.

Table III shows the results for the WAT2016 Japanese-to-Chinese translation subtask with ASPEC-JC corpus. For NMT, perplexity (abbreviated as ppl) is a useful measure of how well the model can predict a reference translation for the given source sentence. The character-level NMT system with the additional character feature of kanji radicals (and input feeding) got 39.65 BLEU points in the test set, which is an improvement of about 7.7 BLEU points over the best system in the Japanese-to-Chinese translation subtask with ASPEC-JP in WAT2016. Further, by adjusting the dropout parameter to 0.3, we got the best results that the perplexity is 3.07 and the BLEU points are 40.61 in the test set. The improvement proves that the character-level NMT system is very effective for the Japanese-to-Chinese translation.

With the same parameters, compared with the character-level NMT with no additional input feature, the perplexity on the development set of the corpus was improved about 0.1 (down), and the BLEU points on the development-test set and the test set of were improved about 0.6 and 0.4 (up), respectively. Further, our best results, compared with the character-level NMT with no additional input feature, the perplexity on the development set of the corpus was improved about 0.6 (down), and the BLEU points in the development-test set and the test set of were improved about 1.5 and 1.4 (up), respectively.

From Table IV, we can see the translation results of the first sentence is gradually improved. The translations of the underlined parts in the sentences show the differences between the translation results of the systems. Our proposed system correctly translated the word 着生 (epiphytic) that the other systems did not translated correctly, and translated the word 必要 (should be) that the other systems did not translate. For the second sentence, the underlined parts show that the word 界面 (surface), which only our proposed system had correctly translated, even the reference Chinese translation was wrong. However, no system could translate the Japanese nouns ヘキサデシルトリメチルアンモニウムブロミド (hexadecyltrimethylammonium bromide) correctly. It is difficult even for humans to translate. The above can prove that the radicals are able to expand the meaning of the characters, and in a sense, could completely translate the correct meaning.

## V. CONCLUSION

In this paper we show that the additional input feature of kanji radicals can be easily provide further improvements for character-level Japanese-Chinese MT. We found that the proposed method improves the model quality. Over the character-level NMT with no additional input feature, the perplexity on the development set of the ASPEC-JC corpus was improved about 0.6 (down), and BLEU points are improved about 1.5 and 1.4 (up) in the development-test set and test set of the corpus, respectively.

TABLE II. A JAPANESE SOURCE SENTENCE AND THE CORRESPONDING RADICALS

| Source sentence | 溝幅は１０ｍｍ以上が必要と推定した。<br>(estimated that the groove width should be 10 mm or more.) |
|---|---|
| Radical features | 水巾水一雨艹艹人一力心西止手宀丿大竹 |

TABLE III. RESULTS

| System | Japanese → Chinese | | |
|---|---|---|---|
| | Ppl ↓ | BLEU ↑ | |
| | dev | devtest | test |
| Baseline (WAT2016 published) | | | 31.98 |
| Character-level NMT | 3.82 | 38.24 | 38.57 |
| Character-level with input feeding | 3.73 | 39.03 | 39.25 |
| Character-level with radical input feature and input feeding | 3.64 | 39.62 | 39.65 |
| Our best results | **3.07** | **40.58** | **40.61** |

---

[1] Han character library for CJKV languages, https://github.com/cburgmer/cjklib

[2] http://github.com/fxsjy/jieba

TABLE IV. TRANSLATION EXAMPLES ILLUSTRATING THE EFFECT OF THE ADDITIONAL INPUT FEATURE OF RADICALS

| | | Input sentences and their translations |
|---|---|---|
| Input Japanese sentence | 1 | 製造工程の作業性や着生状況を解析し，摂食抑制用の溝幅は10mm以上が必要と推定した。<br>(We analyzed the workability of the manufacturing process and the state of epiphytic and estimated that the groove width for eating suppression should be 10 mm or more.) |
| | 2 | 硫酸ジルコニウムメソ多孔質構造体(ZS)は，Zr(SO4)2·4H2O と界面活性剤ヘキサデシルトリメチルアンモニウムブロミドを用いて，100℃で 48 時間水熱反応して合成した。<br>(The zirconium sulfate mesoporous structure (ZS) was synthesized by hydrothermal reaction at 100 °C for 48 hours using Zr(SO4)2·4H2O and the surface active agent hexadecyltrimethylammonium bromide.) |
| Reference Chinese translation | 1 | 解析了制造工程的工作状况及着生状况，推断了抑制摄食时的水沟宽度为10mm以上。 |
| | 2 | 硫酸锆介多孔质构造体(ZS)是使用 Zr(SO4)2·4H2O 和界面活性剂溴化十六烷基三甲铵，在 100℃下经过 48 小时水热反应合成的。 |
| Output of character-level NMT | 1 | 分析了制造工程的作业性和着生状况，进食抑制用的沟幅在10mm以上。 |
| | 2 | 硫酸锆的多孔质结构体(ZS)使用 Zr(SO4)2·4H2O 和界面活性剂己烷基三甲基铵，在 100℃下进行 48 小时水热反应合成。 |
| Output of character-level NMT with input feeding | 1 | 分析了制造工序的工作性和着生状况，并推测出摄食抑制用的沟宽为10mm以上。 |
| | 2 | 硫酸锆的多孔质结构体(ZS)使用 Zr(SO4)2·4H2O 和界面活性剂六甲基三甲酰胺，在 100℃下进行 48 小时水热反应合成。 |
| Output of character-level NMT with radical input feature and input feeding | 1 | 分析制造工序的工作性和附生状况，推定摄食抑制用的沟宽需要10mm以上。 |
| | 2 | 硫酸锆的多孔质结构体(ZS)，使用 Zr(SO4)2·4H2O 和表面活性剂己烷基三甲氨酸溴胺，在 100℃下进行 48 小时水热反应合成。 |
| Output of our best results | 1 | 对制造工序的作业性和附着状况进行分析，推测用于抑制摄食的沟宽需要10mm以上。 |
| | 2 | 硫酸锆膜多孔质结构体(ZS)使用 Zr(SO4)2·4H2O 和表面活性剂十六烷基三甲基溴铵，在 100℃下进行 48 小时的水热反应合成。 |

There is potential to explore the other features for neural MT, which might prove to be even more helpful than the one we investigated, and the feature we investigated may prove especially helpful for some other translation tasks, for example, Chinese-Japanese, Japanese-English, etc.

ACKNOWLEDGMENT

Zhang is supported by the program of China Scholarships Council (No. 201708050078).